\documentclass[preprint,review,12pt]{elsarticle}

\usepackage{amssymb}
\usepackage{framed,multirow}
\usepackage{booktabs}
\usepackage{caption}
\usepackage{subcaption}
\usepackage{makecell}
\usepackage{enumitem}
\usepackage[table]{xcolor}
\usepackage{latexsym}
\usepackage{xurl}
\usepackage{xcolor}
\usepackage{tabularray}

\definecolor{newcolor}{rgb}{.8,.349,.1}

\begin{document}

\begin{frontmatter}



%%\title{Open Set Recognition: Concepts, problems, state of the art, and research directions}

\title{Managing the unknown: a survey on Open Set Recognition and tangential areas}

\author[1,2]{Marcos Barcina-Blanco\corref{cor1}} 
\cortext[cor1]{Corresponding author. Email: marcos.barcina@tecnalia.com}
\author[1]{Jesus L. Lobo}
\author[2]{Pablo Garcia-Bringas}
\author[1,3]{Javier Del Ser}
%% \affiliation[label1]{organization={},%Department and Organization
%%             citysep={}, % use if no comma needed between city and postcode%%             

\affiliation[1]{organization={TECNALIA},
	addressline={Basque Research and Technology Alliance (BRTA)}, 
	city={Derio}, 
	postcode={48160}, 
	country={Spain}}

\affiliation[2]{organization={Faculty of Engineering},
                addressline={University of Deusto},
                city={Bilbao}, 
                postcode={48007}, 
                country={Spain}}

\affiliation[3]{organization={Department of Communications Engineering},
                addressline={University of the Basque Country (UPV/EHU)}, 
                city={Bilbao}, 
                postcode={48013}, 
                country={Spain}}

\begin{abstract}
In real-world scenarios classification models are often required to perform robustly when predicting samples belonging to classes that have not appeared during its training stage. Open Set Recognition addresses this issue by devising models capable of detecting unknown classes from samples arriving during the testing phase, while maintaining a good level of performance in the classification of samples belonging to known classes. This review comprehensively overviews the recent literature related to Open Set Recognition, identifying common practices, limitations, and connections of this field with other machine learning research areas, such as continual learning, out-of-distribution detection, novelty detection, and uncertainty estimation. Our work also uncovers open problems and suggests several research directions that may motivate and articulate future efforts towards more safe Artificial Intelligence methods.
\end{abstract}


\begin{highlights}
    \item Common grounds on the concepts and definitions of Open Set Recognition
    \item Overlaps with continual learning, novelties, uncertainty estimation, and OoD
    \item Recent contributions and methodological approaches to Open Set Recognition
    \item A prospect of existing problems and future research lines for Open Set Recognition
\end{highlights}


\end{frontmatter}



\section{Introduction}
\label{sec1}

Up until this point and persisting into the present, the predominant utilization of Artificial Intelligence has been centered around models capable of executing particular tasks, often operating under careful guidance and complete supervision. Although this method has demonstrated its efficacy in numerous scenarios and remains relevant, there is an undeniable shift towards emphasizing autonomy and broader applicability in open scenarios. Consequently, there is a fervent quest for the emergence of a new era of Machine Learning (ML) models characterized by enhanced autonomy and generalization to perform a wide variety of tasks. But most formulations of such tasks still assume a so-called \emph{closed set} scenario: all samples (or instances) at inference time belong to at least one of the classes existing in the training data from which the ML model was learned. Unfortunately, in many real-world circumstances, this \emph{ closed set} assumption may not necessarily hold, giving rise to \emph{open set} environments where Unknown Classes (UC) can emerge at testing time. When this occurs, the model must detect the emergence of UC; otherwise, ML models designed under the \emph{open set} assumption will incorrectly classify instances belonging to UC as any of the known classes (KC), often with a high confidence in their predictions.

In this context, the Open Set Recognition (OSR) field has emerged \cite{geng2020recent} to address this problem by endowing ML models with the capacity to detect (and adapt) their knowledge to the appearance of new classes. From a technical perspective, OSR departs from a limited number of classes available for learning the ML model, whereas instances belonging to UC may appear during the testing phase. Models are required not only to detect and reject unknown instances, but also to correctly classify samples that belong to the KC. This seminal definition of OSR was stated in \cite{scheirer2012toward}, which also introduced the \textit{over-occupied space} problem. This problem refers to the extra feature space that \emph{open set} classifiers assign to each class to create more meaningful boundaries for the KC. While this strategy is effective in improving the generalization ability of the ML model under a \emph{open set} assumption, it leads to incorrect classification of UC in open settings. Since then, several OSR methodologies have been proposed over the years to cope with the aforementioned problem, from the modification of traditional ML algorithms to the development of Deep Learning models capable of autonomously detecting UC. More recently, attempts to combine clustering and classification algorithms \cite{coletta_combining_2019, wang_open-set_2022} have shown promising results. Although the detection of UC has been central for the research community working in OSR, other areas in ML research have gravitated on problems that relate closely to the \emph{open set} assumption underneath OSR, such as Continual Learning (CL) \cite{gunasekara_survey_2023}, Novelty Detection (ND), Out-of-Distribution (OoD) detection, or Uncertainty Estimation (UE).

From a practical standpoint, OSR has been proven beneficial to identify the unknown in practical applications prone to the appearance of unseen classes, including the fields of computer vision \cite{mazur_feature-realistic_2023,sisti_open-set_2022,zhao2023open,saranrittichai2022multi} and natural language processing \cite{zheng_out--domain_2022,liu_towards_2023}, or sub-fields such as cybersecurity \cite{soltani_adaptable_2023}, biometric recognition \cite{shao_towards_2022, liu_sphereface_2023}, face recognition \cite{liu_sphereface_2023}, or monitoring systems \cite{li2023importance}. It is worth mentioning the Open World Learning (OWL) paradigm \cite{parmar_open-world_2022}, which can be regarded as an extension of OSR, where the model has not only to detect UC, but also to characterize and to incorporate UC to the model.

This work also finds its motivation in: 
\begin{itemize}
    \item establishing common grounds on the concepts and definitions managed in OSR, which have been evolving since the last few years,
    \item clarifying the similarities and differences with the aforementioned tangential areas, showing the manifold points of friction with them, such as CL, ND, OoD detection, or UE,
    \item the upsurge of contributions that tackle the OSR problem with different methodological approaches that have been reported in recent times, and
    \item concluding with a prospect of existing problems and research lines of interest for the advance in this field
\end{itemize}

This work is organized as follows: Section 2 delves into the OSR concept, problem, and its related areas, while Section 3 provides a literature review; the Section 4 presents the challenges and future work for OSR scenarios; finally, Section 5 draws the conclusions related to this survey.

\section{OSR: concepts and related research areas}

We begin our survey by briefly looking at the mathematical statement of the OSR problem and several definitions needed for its proper understanding (Section \ref{ssec:definition}), including the \textit{over-occupied space} problem mentioned previously. We then discuss research areas that are closely related to OSR or overlap with each other to some extent. Section \ref{ssec:areas} details what such areas share in common, what they differ in, and review existing approaches that address scenarios that fall within the \emph{open set} assumption.

\subsection{OSR concepts and problem definition} \label{ssec:definition}

\begin{figure}[ht]
	\centering
	\begin{subfigure}[b]{0.45\textwidth}
		\centering
		\includegraphics[width=\textwidth]{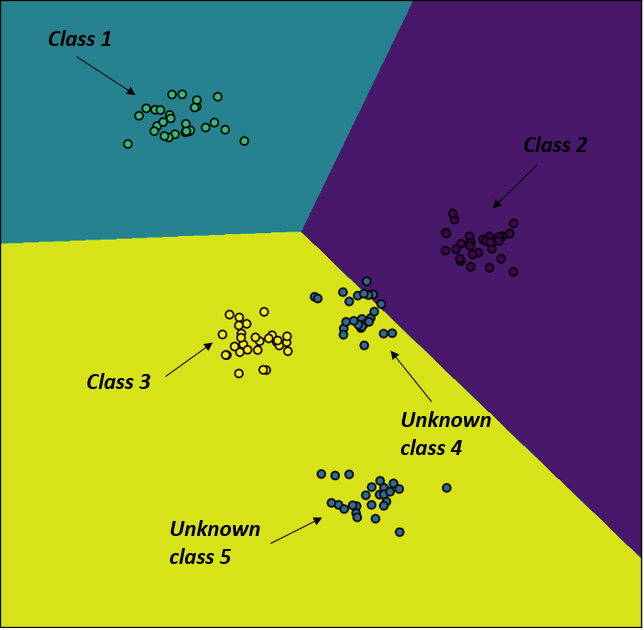}
		\caption{Closed set classifier}
		\label{fig:closed-set classification}
	\end{subfigure}
	\hfill
	\begin{subfigure}[b]{0.45\textwidth}
		\centering
		\includegraphics[width=\textwidth]{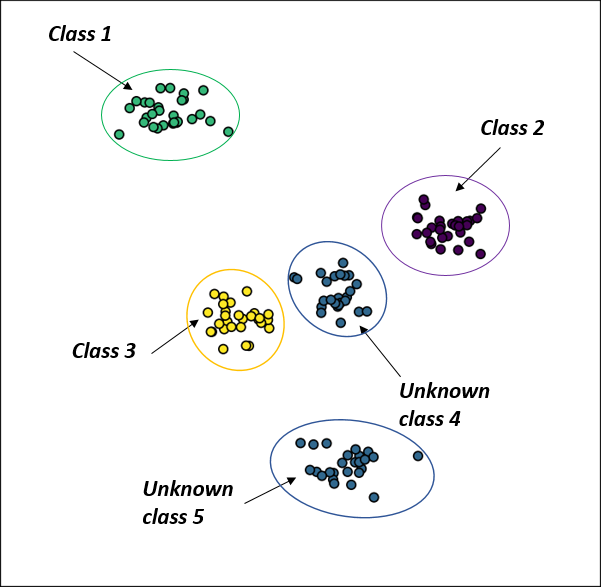}
		\caption{Open Set classifier}
		\label{fig:osr}
	\end{subfigure}
	\caption{Differences between \emph{open set} and \emph{open set} classification models. (a) Shows the decision boundaries of an \emph{open set} classifier for a problem with $N=3$ KC and $\Omega=2$ UC. All the feature space is divided between the KC, so that instances from UC arriving at inference time will be incorrectly classified, in some cases (e.g. unknown class 4) with high class probability. (b) It depicts that \emph{open set} classifiers delimit the feature space that each KC occupies, allowing for the effective detection of UC.}
	\label{fig:osr comparison}
\end{figure}

Conventional classification tasks formulated under a \emph{open set} assumption assume that all samples belong to one of the KC during training. As such, a classification model $M$ is trained over a dataset comprising KC $\mathcal{C}_{KC}=\{Y_1, Y_2, \ldots, Y_N\}$. Once learned, the model is queried with instances at inference time, which produces a predicted class belonging to $\mathcal{C}_{KC}$ for each test instance. In this \emph{open set} scenario, given any input $\mathbf{x}$, the classifier $M$ estimates a probability distribution over all training classes $p(y|\mathbf{x})$, with $y \in \mathcal{C}_{KC}$. This will yield a classification error whenever the classifier is queried with a new instance $\mathbf{x}$ that does not belong to any of the classes in $\mathcal{C}$. Figure \ref{fig:closed-set classification} schematically represents this problem for a problem with three KC ($|\mathcal{C}_{KC}|=N=3$) and 2 features ($|\mathbf{x}|=2$).\par

As shown in Figure \ref{fig:osr comparison}, traditional \emph{open set} classifiers assign the entire feature space to the $N$ KC in order to create more meaningful discriminatory boundaries between KC and ultimately, attain a better generalization under the \emph{open set} assumption. However, this over-occupation of the space means that any unknown sample will fall in this space and be incorrectly classified as one of the KC. 

OSR undertakes this problem by devising a scenario in which knowledge of the full set of classes is not available during training. An \emph{open set} classification model is trained to discriminate among KC in $\mathcal{C}$, but test inputs may also come from UC $\mathcal{C}_{UC}$. In this situation, the model must be capable of predicting classes from $\mathcal{C}_{KC} \:\cup\: C_{UC}  = \{Y_1, Y_2, \ldots, Y_N, Y_{N+1}, \ldots, Y_{N+\Omega}\}$, where $\mathcal{C}_{UC}$ represent all the UC not seen during training. The model should be able to predict any new instance $\mathbf{x}$ as either belonging to one of the KC in $C_{KC}$ or classify it as unknown (i.e., as belonging to $C_{UC}$). 

Fig. \ref{fig:osr comparison} also shows all the feature space that is far from the training instances belonging to KC. This is known as the \emph{Open Space}. The work in \cite{scheirer2012toward} discusses on the risk associated to labeling instances that are far from the training samples in the Open Space, giving rise to the \emph{Open Space Risk} $R_O(f)$ defined as:
\begin{equation}
R_O(f) = \frac{\int_O f(\mathbf{x})d\mathbf{x}}{\int_{S_O} f(\mathbf{x})d\mathbf{x}},
\end{equation}
which is a relative measure of positively labeled Open Space $O$ to the total measure space $S_O$, covering the known positive samples and $O$. The function $f$ is a recognition function that depends on the OSR model/strategy in use, where $f(\mathbf{x})=1$ means that a known class is recognized ($f(\mathbf{x})=0$ otherwise). The more open space labeled as positive, the greater $R_O$ becomes, and hence the higher the risk taken by the model when operating in an \emph{open set} scenario will be.

\subsection{OSR in other relevant research areas} \label{ssec:areas}

The problem stated in Section \ref{ssec:definition} is close to the focus pursued in other areas in ML research, which are discussed next.

\subsubsection{Novelty Detection} \label{ssec:areas:ND}
This area focuses on the detection of new test samples that have not been seen during training. Novelty Detection (ND) should not be confused with Anomaly Detection (AD). While AD assumes that abnormal samples may already exist in the training data and focuses on finding them, ND assumes clean training data without any anomalies and expects a semantic shift (appearance of novel classes) to occur during the testing phase. Clearly, ND shares many similarities with OSR, but there are still several subtle differences. To begin with, ND is generally considered an unsupervised task, while in OSR the model has access to the labels of the KC. This is consequent with their own respective goals. Unlike OSR, ND focuses only on detecting novel samples, classification of samples belonging to the KC is not required. Although ND tasks may also be formulated in multi-class settings, their ultimate goal is the detection of novel samples, i.e. a binary classification problem between KC and UC. In fact, the use of one-class models has become prevalent in this area \cite{yerima_semi-supervised_2022} due to the inherently unsupervised nature of ND. Other recent advances in ND employ autoencoder neural network architectures \cite{salehi_arae_2020, lo_adversarially_2023, huang_calibrated_2023}, GAN architectures \cite{xia_gan-based_2022}, diffusion models \cite{mirzaei_fake_2022}, or three-way clustering \cite{shah_three-way_2021}.

\subsubsection{Continual Learning} \label{ssec:areas:CL}

Continual learning (CL), often denoted as continuous learning or lifelong (machine) learning, focuses on a knowledge-driven paradigm, where the knowledge acquired in the past is retained and used to learn new things with little data or less effort \cite{chen2018lifelong,wang2023comprehensive}. Under this paradigm we usually find related learning paradigms such as online learning, incremental learning, or stream learning; in essence, all of them are associated with the gradual provision of data throughout their lifetime. Despite in many cases the distinction between these terms is often not strictly defined, nuances aside, they face the challenge of learning from data that are continuously generated over time (data stream). This scenario is also prone to encountering novel samples during the testing phase, especially when the statistical distribution of the data stream is not stationary. Since OSR pursues to detect when a model faces an unknown sample, it is natural that some research has been done under streaming conditions, where identifying new classes from the group of unknown samples and updating the model with the new knowledge are added challenges \cite{gao_sim_2019,leo_moving_2020}. There are few works that explicitly reference the problem of OSR in CL \cite{mundt2022unified}, and over streaming data due to the overlap between OSR and the concept evolution field of stream learning \cite{bouguelia_adaptive_2016,mohamad_bi-criteria_2018,zhang_knnens_2022}.

\subsubsection{Out of Distribution Detection} \label{ssec:areas:OoD}
This area is also closely related to OSR. It focuses on detecting and dealing with Out-of-Distribution (OoD) samples at testing time, i.e. samples from UC. OoD features small differences when compared to OSR. First, some OoD methods involve exposing the model to a number of out-of-distribution instances during the training process to help models learn to discriminate between in-distribution and out-of-distribution data. The use of these training OoD data clashes with the assumptions made in OSR. Another difference arises in the benchmarking protocol followed by studies related to these two research areas: while OSR models usually split a single dataset into partitions with instances belonging to known (in-distribution) and unknown (out-of-distribution) classes, OoD models consider an entire dataset as the in-distribution, whereas several different datasets are utilized as out-of-distribution data. All in all, OoD detection techniques can be used to perform OSR in multi-class settings. The detection of OoD samples has been approached from different means in the literature \cite{yang2021generalized}, even comparing it to OSR \cite{gillert2021towards}. Among them, it is worth noting the energy-based models \cite{song_how_2021,liu_energy-based_2021, elflein_masters_2023,wu_energy-based_2023}, the distance-based models \cite{sun_out--distribution_2022}, the density-based approaches \cite{zisselman_deep_2020}, or the diffusion approaches \cite{wu2023deep}.

\subsubsection{Uncertainty estimation} \label{ssec:areas:UE}
The Uncertainty Estimation (UE) provides the ability to estimate and quantify the uncertainty of individual predictions. Unlike aleatory uncertainty that is attributed to the inherent random processes in nature, epistemic uncertainty is often due to a lack of training data, and thus it is particularly interesting in OSR for quantifying the knowledge uncertainty and rejecting the OoD inputs \cite{mundt2019open,pires2020towards}. 

\section{OSR literature review}

There have been many works done in OSR from different modeling perspectives. Previous surveys on OSR \cite{scheirer2014probability,geng2020recent,mahdavi_survey_2021,vaze_open-set_2022} agree on the categorization of existing approaches in two main categories: \emph{discriminative} and \emph{generative}, although we can find some approximations that combine the two \cite{perera2020generative}. Discriminative methods model the existing KC data into tighter spaces, while generative approaches synthesize fake UC instances to train the model. This taxonomy for representative OSR works and its closest research areas are shown in Table \ref{tabosrworks}. 
We now proceed by examining contributions classified in the taxonomy, including those proposing discriminative approaches (Section \ref{sec:desc}) and generative methods (Section \ref{sec:gen}). Next, we pause at discriminative approaches that combine a clustering method and a classification model (Section \ref{sec:clustclass}), giving it its own subsection (\ref{sec:clustclass}); this alternative to the modification of classification models to reduce the over-occupied feature space is central in our prospects about the future of the OSR area offered later in Section \ref{challs_bey}.

% \usepackage{multirow}
\begin{table}[]
\caption{Overview of the works on OSR, applications, and related areas.}
\label{tabosrworks}
\begin{tabular}{|c|cc|c|}
\hline
\textbf{Field} &
  \multicolumn{2}{c|}{\textbf{Approaches}} &
  \textbf{References} \\ \hline
\multirow{10}{*}{\begin{tabular}[c]{@{}c@{}}Open Set\\ Recognition\end{tabular}} &
  \multicolumn{2}{c|}{Surveys} & \cite{geng2020recent, mahdavi_survey_2021, vaze_open-set_2022,fahy_scarcity_2022, parmar_open-world_2022}
   \\ \cline{2-4} 
 &
  \multicolumn{1}{c|}{\multirow{6}{*}{\begin{tabular}[c]{@{}c@{}}Discriminative\\ methods\end{tabular}}} &
  SVM & \cite{scheirer2012toward, ma2018towards}
   \\ \cline{3-4} 
 & \multicolumn{1}{c|}{} & EVM       & \cite{henrydoss_incremental_2017, rudd2017extreme, henrydoss_enhancing_2020} \\ \cline{3-4} 
 & \multicolumn{1}{c|}{} & Distance  & \cite{bendale_towards_2015, mendes_junior_nearest_2017, hui_new_2022,cardoso2017weightless} \\ \cline{3-4} 
 & \multicolumn{1}{c|}{} & Prototype & \cite{yang_convolutional_2022, chen_learning_2020, lu_pmal_2022, liu_learning_2023, liu_towards_2023} \\ \cline{3-4} 
 & \multicolumn{1}{c|}{} & DNN       & \makecell{\cite{bendale_towards_2016, shu_doc_2017, shu_unseen_2018, oza2019c2ae, gao_sim_2019, leo_moving_2020} \\\cite{shao_towards_2022,soltani_adaptable_2023,liu_sphereface_2023,mandivarapu2022deep,vendramini2021opening, komorniczak2024torchosr}} \\ \cline{3-4} 
 &
  \multicolumn{1}{c|}{} &
  \begin{tabular}[c]{@{}c@{}}Clustering-\\Classification\end{tabular} & \cite{coletta_combining_2019, zhang_hybrid_2020, wang_open-set_2022}
   \\ \cline{2-4} 
 &
  \multicolumn{1}{c|}{\multirow{2}{*}{\begin{tabular}[c]{@{}c@{}}Generative \\ methods\end{tabular}}} &
  GANs & \makecell{\cite{ge_generative_2017, jo_open_2018, neal_open_2018,kong_opengan_2021}\\\cite{goodman_generative_2022,sisti_open-set_2022,pal_morgan_2023, xia_adversarial_2023,engelbrecht2023link}}
   \\ \cline{3-4} 
 & \multicolumn{1}{c|}{} & Diffusion & \cite{mirzaei_fake_2022, wu2023deep} \\ \cline{2-4} 
 & \multicolumn{2}{c|}{Applications}       & \makecell{\cite{mazur_feature-realistic_2023,sisti_open-set_2022,zhao2023open,saranrittichai2022multi,zheng_out--domain_2022, liu_towards_2023}\\\cite{soltani_adaptable_2023,shao_towards_2022, liu_sphereface_2023,li2023importance, lopez-lopez_incremental_2022}} \\ \hline
\multirow{4}{*}{\begin{tabular}[c]{@{}c@{}}Related \\ areas\end{tabular}} &
  \multicolumn{2}{c|}{Novelty detection} & \makecell{\cite{yerima_semi-supervised_2022,salehi_arae_2020,lo_adversarially_2023}\\\cite{huang_calibrated_2023,xia_gan-based_2022,mirzaei_fake_2022,shah_three-way_2021}}
   \\ \cline{2-4} 
 & \multicolumn{2}{c|}{Continual learning} & \makecell{\cite{chen2018lifelong,wang2023comprehensive,gao_sim_2019,leo_moving_2020}\\\cite{bouguelia_adaptive_2016,mohamad_bi-criteria_2018, zhang_knnens_2022}} \\ \cline{2-4} 
 & \multicolumn{2}{c|}{OoD detection}      & \makecell{\cite{yang2021generalized, gillert2021towards, song_how_2021,liu_energy-based_2021}\\\cite{elflein_masters_2023, wu_energy-based_2023,sun_out--distribution_2022,zisselman_deep_2020, wu2023deep}} \\ \cline{2-4} 
 &
  \multicolumn{2}{c|}{Uncertainty estimation} & \cite{mundt2019open,pires2020towards}\\ \hline
\end{tabular}
\end{table}

\subsection{Discriminative approaches} \label{sec:desc}

As stated above, discriminative approaches focus on reducing the \emph{Open Space Risk} ($R_O$) by modeling the data into smaller spaces in the feature space. This has been done mainly by adapting traditional ML methods \cite{scheirer2012toward, ma2018towards, rudd2017extreme, bendale_towards_2015, mendes_junior_nearest_2017, hui_new_2022, yang_convolutional_2022, chen_learning_2020, lu_pmal_2022, liu_learning_2023,boudiaf2023open}, or by tailoring the neural network training process to perform inference in environments subject to the appearance of new concepts \cite{nguyen_deep_2015, bendale_towards_2016, shu_doc_2017, shu_unseen_2018, cevikalp_anomaly_2023}. 

The authors in \cite{scheirer2012toward} explained the OSR problem as a balance between reducing $R_O$ and the empirical risk. For this purpose, they propose a so-called ``1-vs-set Machine'' model based on Support Vector Machines (SVM). In \cite{ma2018towards} a framework that combines a Conditional Random Field and a Probability of Inclusion SVM was proposed for the rejection of UC. The authors in \cite{rudd2017extreme} presented the Extreme Value Machine (EVM), a model based on statistical Extreme Value Theory (EVT), where this approach naturally reduces $R_O$ and decides which new samples are unknown based on a threshold.

Distance-based models have also been modified in order to work in an \emph{open set} scenario. One of the first works contributed in this direction was the Nearest Non Outlier (NNO) algorithm, a modification of the Nearest Class Mean classifier \cite{bendale_towards_2015}. Furthermore, other traditional distance-based classifiers have been adapted in different ways, such as the Nearest Neighbor Classifier \cite{mendes_junior_nearest_2017, hui_new_2022}. In \cite{mendes_junior_nearest_2017} an extension to the k-nearest neighbors (k-NN) algorithm was proposed, called Nearest Neighbor Distance Ratio (NNDR). More recently, the work in \cite{hui_new_2022} assumes that samples from KC (UC) are likely to be surrounded by other samples belonging to KC (UC). The Otsu’s method \cite{liu_otsu_2009} is used to select the optimal threshold to classify into KC and UC; after the classification is done, samples considered to be KC are passed to a Random Forest Classifier learned from the training samples to perform regular \emph{open set} classification. In \cite{cardoso2017weightless} authors presented a weightless artificial neural network model to achieve an effective combination of classification with precise identification of extraneous data.

In recent years, Prototype Learning has elicited promising results in the OSR field. The use of prototypes allows for more compact feature representations of classes, which naturally creates clearer boundaries between KC and UC in \emph{open set} scenarios. In \cite{yang_convolutional_2022} authors introduced a Convolutional Prototype Network (CPN) where prototypes per class are jointly learned during training. The authors in \cite{chen_learning_2020} not only used prototypes to represent classes, but also learned discriminative reciprocal points for them. These reciprocal points can be understood as the opposite of the prototype, i.e., what is not the class under target. They help to explore the open space, naturally reducing the open space risk $R_O$. Similarly, the authors in \cite{lu_pmal_2022} propose that implicitly learned prototypes tend to create undesired prototypes from low-quality samples, and show redundancy in similar prototypes of one category. Differently from the previous works, the work in \cite{liu_learning_2023} resorts to Gaussian distributions to represent prototypes instead of sets of feature vectors; they argue that the distribution of KC in the latent feature space can be represented by one or several Gaussian-distributed prototypes.

Deep Neural Networks (DNN) offer powerful representation abilities that can be beneficial for any ML task. DNN used for classification task usually include a last softmax layer to produce a probability distribution over the KC, posing a problem in OSR settings. Furthermore, DNNs often produce high confidence scores for samples belonging to UC \cite{nguyen_deep_2015}. Approaches that rely on thresholding softmax probability scores and assuming that UC inputs produce low probability for all classes have proven to not reach optimal solutions in OSR \cite{bendale_towards_2016}. Several workarounds have been explored to overcome this issue, such as replacing the Softmax function with OpenMax \cite{bendale_towards_2016}. Other proposals \cite{shu_doc_2017,shu_unseen_2018} replace the softmax function with a ``1-vs-rest'' layer based on sigmoid functions, which is able to naturally reject unknown samples. In \cite{oza2019c2ae} an OSR algorithm using class conditional auto-encoders was proposed in which the encoder learned \emph{ closed set} classification while the decoder learned to identify KC and UC. The authors in \cite{cevikalp_anomaly_2023} proposed deep compact hypersphere classifiers for OSR. They consider the hypersphere centers as learnable parameters, and update them based on the features of new samples. Other approaches have formulated active learning as an OSR problem \cite{mandivarapu2022deep}, or have used generative information \cite{vendramini2021opening}. Finally, an OSR python module that facilitates the correct experimental evaluation of DNN-based solutions has also been developed very recently \cite{komorniczak2024torchosr}.

\subsection{Generative approaches} \label{sec:gen}

Unlike their discriminative counterparts, generative methods seek to produce synthetic data samples that represent the UC, so that the model is trained over real and synthetic samples. Generative Adversarial Networks (GAN) were first applied to OSR \cite{ge_generative_2017}. These models generate synthetic samples and assign them a new label that represents the UC, so that, when added to the training data from which the classifier is learned, allows predicting samples belonging to UC. Other attempts reported in \cite{jo_open_2018, kong_opengan_2021} also synthesize instances that are representative of UC, showing that, when used to train the model, make it more robust against samples from actual UC. Similarly, the authors in \cite{neal_open_2018} generate counterfactual images, i.e. those similar to real instances from the KC that are close to the UC. The DPAA algorithm proposed in \cite{goodman_generative_2022} generates synthetic anomalies close to the \emph{open set} boundary of the classifier; the distance at which samples are generated is balanced against a constraint that ensures that the \emph{open set} classification performance is not degraded. Synthetic samples generated in this manner are used to reduce $R_O$ while not increasing the empirical risk. The authors in \cite{pal_morgan_2023} proposed to use two GANs: generated adversarial samples that have low noise variance are used to augment the density of KC, while samples with high noise variance are dispersed into the open space in order to decrease $R_O$. Adversarial samples have also been used to enhance other discriminative methods such as prototype learning \cite{xia_adversarial_2023}. In the case of \cite{engelbrecht2023link}, authors studied the relations between semi-supervised learning and OSR under the context of GANs.
As can be inferred from the reviewed literature, the key to generative methods is to synthesize representative samples of all possible UC, and thus maximize the decrease of $R_O$ resulting from the inclusion of such samples in the training data of the classifier.

\subsection{Clustering-classification discriminative approaches} \label{sec:clustclass}

As we have seen, former discriminative attempts to solve this problem have tried to directly modify the classifiers so that they are able to deal with inputs belonging to UC. Alternatively, clustering models can be combined with classification to address this problem \cite{geng2020recent}. Clustering models depart from a measure of similarity among instances, so that those belonging to a certain KC are closer to each other than to those belonging to the rest of KC and UC. They naturally partition the feature space when modeling KC. In order to deal with the \textit{over-occupied space} problem, clustering algorithms can help better adjust KC to a smaller area, while classification models are used to discriminate between them.

Many efforts for combining clustering and classification exist for \emph{open set} scenarios, which can be implemented either sequentially or simultaneously. Those which work in a sequential fashion first apply clustering techniques over the training data so as to characterize the feature space corresponding to the KC. This characterization is then exploited to improve the robustness of the subsequent classification model against data belonging to UC. A common approach following this sequential procedure can be found in \cite{samunnisa_intrusion_2023}, where both techniques are combined for malicious attacks classification. Another proposal using clustering to improve classification is \cite{henrydoss_enhancing_2020}, which first applies DBSCAN over the training data, and uses the obtained centroids as the Extreme Vectors of an EVM, hence overriding the need for comparing all training samples and significantly reducing the computational effort of the overall classifier. In \cite{wang_open-set_2022}, images belonging to each class of camera models are first clustered; then a SVDD (Support Vector Data Description) model is trained over each of the clusters to create several hyperspheres. Instances that fall outside of their modeled space are considered to belong to UC. The rationale behind this approach is that several hyperspheres adapt better to the distribution of each class and thus reduce the \textit{over-occupied space}. While these methods result in better performance, they do not take full advantage of the combination, since the clustering techniques do not benefit from the information produced in the classification stage \cite{qian_simultaneous_2012}.

Techniques simultaneous applying clustering and classification aim to ensure that both algorithms receive feedback from each other during the training phase. Early contributions in this line proposed to combine both criteria into a single objective of the model, which was then optimized \cite{cai_simultaneous_2009,qian_simultaneous_2012}. Although there has been far more research in this line of work \cite{bharill_improved_2011, liu_particle_2014, luo_learning_2016, li_learning_2020}, all of them assume a \emph{open set} scenario. These works share that clustering and optimization depend on the cluster centers, and use a relation matrix in order to compute the relationship between cluster membership and class membership. Then, each of them applies its own multi-objective optimization framework to find the matrix and centers that best satisfy the clustering and classification indexes. 

Other simultaneous approaches have combined ensembles of classifiers and clustering models \cite{acharya_c3e_2011}. The class probability distribution for each test instance (from the classifiers ensemble) and a similarity matrix (from the clusters ensemble) are combined together to yield a posterior class probability assignment for each test instance. In \cite{coletta_combining_2019} this work was extended to make it more suitable for open world settings. While this approach rendered a good performance in OSR, it still undergoes issues when the space of new UC overlaps with the space of KC, and requires significant processing time to retrain the model after each iteration. Further, the work in \cite{zhang_hybrid_2020} combines a discriminative classifier and a flow-based model to address OSR. Flow models can predict the probability density for each sample: samples that have a large probability density are likely to belong to the training distribution, whereas instances that have a small probability density value usually belong to an UC. Instead of adding a classifier after the flow model, a joint embedding is created for both the classifier and the flow model. This is done in order to ensure more discriminative expressiveness in the embedding space (rather than using the space that the flow model creates on its own).

To the best of our knowledge, beyond the works reviewed above very scarce attempts have been done to combine clustering and classification towards reducing the \textit{over-occupied space} from an OSR perspective. Our prospects offered later in this manuscript will further revolve around this identified niche.

\section{OSR challenges and beyond}\label{challs_bey}

In what follows we offer our view on the challenges that this area faces, and outline research directions that can be pursued to address them efficiently:

\paragraph{Open space risk} Following up its definition in Section 2, our literature review has revealed that, despite the activity noted in the area during the last year, reducing the open space risk remains an open problem. Decision boundaries delineated by off-the-shelf ML classifiers tend to over-occupy the feature space for the KC. Managing the space out of the feature space regions corresponding to the KC is still a hard task that increases its difficulty with the openness of the problem \cite{scheirer2012toward}. Although the \textit{over-occupied space} is a common issue for all perspectives that deal with UC, it has never been technically approached from the perspective of concept evolution and OoD detection. We envision that there can be profitable synergies between these two close research areas and OSR that can endow ML models with an improved robustness against UC. 

\paragraph{Use of thresholds} Many OSR, concept evolution in CL and OoD models rely on a threshold for distinguishing between KC or UC. Computing an optimal threshold is not straightforward and can be time consuming \cite{hui_new_2022}. Furthermore, these thresholds can vary greatly depending on the openness of the problem and the a priori information about the characteristics of the UC that the model will encounter when running inference. Even a threshold for the same set of classes can yield a high open space risk if the data distribution changes and the threshold is not updated accordingly. While an optimal threshold can be dynamically set for each task or set of data, it can still become a problem in an evolving data stream, where new UC may arrive regularly and the model continuously produces predictions for the instances arriving in the stream. Self-adaptive parameter tuning methods \cite{9679910}, including dynamic evolutionary algorithms or reinforcement learning agents, can be an interesting research path to follow in order to achieve OSR techniques capable of operating in highly varying open settings.

\paragraph{Combining clustering and classification} As evidenced by our literature review, most OSR approaches relying on clustering and classification combine both techniques in a sequential way \cite{henrydoss_enhancing_2020, wang_open-set_2022}. This practice has two inherent limitations: on one hand, the clustering algorithm does not benefit from the information provided by the classifier, even though feedback such as the uncertainty about its prediction can be a valuable input for determining whether the test instance is unknown. Very scarce works ensure that both processes benefit from each other in an \emph{open set} scenario. New strategies to hybridize clustering and classification should be targeted by the community in the near future, such as clustering over the space of internal activations when dealing with neural network classifiers \cite{martinez2023novel}.  

\paragraph{Detecting unknown samples and identifying new classes over time} Performing classification of each new sample in isolation is an standard evaluation protocol in OSR \cite{vaze_open-set_2022}. Therefore, instances that arrive before are not taken into account in the decision. However, it is often the case that UC arise gradually over time, imprinting a correlation between successive instances that can be modeled for the sake of a more precise detection of its membership to an KC or UC. This need to account for this time correlation is of utmost importance when dealing with scenarios with intermittently appearing UC: in this case, allowing the OSR technique to memorize long-term relationships between instances detected as UC can be crucial to expedite the UC identification when one of such UC reappears, and/or to support a preemptive reservation of the \textit{over-occupied space} as per the emergence/disappearance dynamics of UC. Data distillation and neural embeddings can provide a low-dimensional representation of samples detected to belong to UC, so that the relationship between unknown concepts over time can be learned and eventually exploited.

\paragraph{Disentangling complex class distributions} A major fraction of the contributions on OSR apply clustering to data and identify new classes from the obtained clusters \cite{gao_sim_2019, leo_moving_2020}. This naive approach does not perform well when known and unknown classes overlap, or when having underpopulated known classes that do not provide enough statistical support to decide whether a new instance belongs to them. This is often the case of OSR in ML tasks formulated over real-world image data subject to a strong content variability (i.e., low domain specificity). We note a lack of reliable strategies for identifying novel classes from unknown samples that perform well over these challenging known class distributions. In this regard we foresee that any means to augment the knowledge of the model about newly emerging classes, including domain-specific meta-information, can help the model identify such classes more reliably.

\paragraph{Updating the model} After UC have been discovered, the model needs to be updated with the new knowledge. Completely re-training the model \cite{gao_sim_2019, leo_moving_2020} may not be a viable strategy in environments with tight time constraints. Similarly, human annotation of new classes is not feasible in most practical problems. Incremental learning strategies such as the update and maintenance of micro-clusters \cite{liao_novel_2023} have recently shown to be promising in stream learning, but have not been extensively researched in OSR and OWL. Likewise, such model updating techniques can be combined with imbalance learning or -- when interaction with the data source is possible -- active learning strategies to enrich the learning process with diverse instances of the newly discovered UC. Finally, well-known issues in continual learning settings, including catastrophic forgetting, also hold in OSR. As such, updating the model with new classes while discriminating known concepts should be also given attention in the future, especially under periodic occurrence patterns.

\section{Concluding remarks}

ML research field is in the search for robust models with a higher level of autonomy and able to reliably manage open environments in which the unknown may emerge. OSR models have gained popularity due to the need for robust classifiers that are able to perform reliably in the open world, dealing with unknown situations that were not seen by the models during its training phase. This review has examined the current state of OSR and the different strategies followed in the literature to tackle the open space risk or \textit{over-occupied space} problem. We have also brought attention to several areas that are closely linked to OSR, such as continual learning, out-of-distribution detection, novelty detection, and uncertainty estimation. Based on the findings of our literature analysis, we have discussed several challenges which, as per our examination of the state of the art, remain unaddressed in the OSR community. Upscaling the current approaches to complex known class distributions, improving the incremental assimilation of newly discovered classes by the model in scarce data regimes, or the exploitation of the temporal correlation between test instances when detecting and characterizing new concepts, are among the ones that should capture efforts in the near future. We hope that this survey becomes a motivating landmark for the research community to join forces and delve deeper into the OSR/OWL research areas.

\section*{Declaration of competing interest}
The authors declare that they have no known competing financial interests or personal relationships that could have appeared to influence the work reported in this paper.

\section*{CRediT authorship contribution statement}
\textbf{Marcos Barcina-Blanco:} Conceptualization, Investigation, Writing - Original Draft, Writing - Review \& Editing, and Visualization. \textbf{Jesus L. Lobo:} Conceptualization, Methodology, Resources, Writing - Review \& Editing, Visualization, and Supervision. \textbf{Pablo Garcia-Bringas:} Conceptualization, Methodology, Writing - Review \& Editing, Project administration, and Funding acquisition. \textbf{Javier Del Ser:} Conceptualization, Methodology, Resources, Writing - Review \& Editing, Project administration, and Funding acquisition.

\section*{Acknowledgements}
All authors approved the version of the manuscript to be published. This research has received funding from the Basque Government (BEREZ-IA project with grant number KK-2023/00012), and the research groups MATHMODE (IT1456-22) and D4K-Deusto for Knowledge (IT1528-22).

 \bibliographystyle{elsarticle-num} 
 \bibliography{cas-refs}





% \end{thebibliography}
\end{document}